\documentclass[letterpaper, 10 pt, conference]{IEEEtran} 

\IEEEoverridecommandlockouts
\usepackage{cite}
\usepackage{balance}
\usepackage{comment}
\usepackage{amsmath,amssymb,amsfonts}
\usepackage{algorithmic}
\usepackage{graphicx}
\usepackage{textcomp}
\usepackage{xcolor}
\usepackage{pgfplots}
\usepackage{tikz}
\usepackage{tikz-3dplot}
\usepackage{url}
\usepackage{hyperref}
\def\BibTeX{{\rm B\kern-.05em{\sc i\kern-.025em b}\kern-.08em
    T\kern-.1667em\lower.7ex\hbox{E}\kern-.125emX}}
\pgfplotsset{compat=1.16}

\hypersetup{
    colorlinks=true,
    linkcolor=blue,
    filecolor=magenta,      
    urlcolor=cyan,
}
    
\begin{document}
\title{\textit {AcinoSet}: A 3D Pose Estimation Dataset \\and Baseline Models for Cheetahs in the Wild}

\author{\IEEEauthorblockN{Daniel Joska$^{1}$, Liam Clark$^{1}$, Naoya Muramatsu$^{2}$, Ricardo Jericevich$^{1}$, Fred Nicolls$^{1}$,
Alexander Mathis$^{3}$, \\ Mackenzie W. Mathis$^{3}$, Amir Patel$^{1}$~\IEEEmembership{Member,~IEEE}}%

\thanks{Research supported by South African National Research Foundation (Grant No. 117744). $^{1}$ African Robotics Unit (ARU), University of Cape Town, South Africa $^{2}$ University of Tsukuba, Japan $^{3}$ 
École Polytechnique Fédérale de Lausanne, Switzerland $^{4}$ 
Corresponding authors: amir.patel@uct.ac.za, mackenzie.mathis@epfl.ch}

}

\maketitle

\begin{abstract}
Animals are capable of extreme agility, yet understanding their complex dynamics, which have ecological, biomechanical and evolutionary implications, remains challenging. Being able to study this incredible agility will be critical for the development of next-generation autonomous legged robots. In particular, the cheetah (\textit{acinonyx jubatus}) is supremely fast and maneuverable, yet quantifying its whole-body 3D kinematic data during locomotion in the wild remains a challenge, even with new deep learning-based methods.  In this work we present an extensive dataset of free-running cheetahs in the wild, called \textit{AcinoSet}, that contains $119,490$ frames of multi-view synchronized high-speed video footage, camera calibration files and $7,588$ human-annotated frames. We utilize markerless animal pose estimation to provide 2D keypoints. Then, we use three methods that serve as strong baselines for 3D pose estimation tool development: traditional sparse bundle adjustment, an Extended Kalman Filter, and a trajectory optimization-based method we call Full Trajectory Estimation. The resulting 3D trajectories, human-checked 3D ground truth, and an interactive tool to inspect the data is also provided. We believe this dataset will be useful for a diverse range of fields such as ecology, neuroscience, robotics, biomechanics as well as computer vision. Code and data can be found at: \url{https://github.com/African-Robotics-Unit/AcinoSet}.

\end{abstract}

\section{Introduction}
The ability to swiftly and robustly maneuver in the world is paramount to survival for many animals. Leveraging the ability of cheetahs---the fastest land mammal---will be useful to build better legged robots, akin to what has been achieved with flying robots\cite{jafferis2019untethered}). Maneuverability also presents interesting case studies on how animals have negotiated trade-offs amongst competing requirements such as safety, economy, stability, robustness and agility~\cite{daley2016non}. The cheetah (\textit{acinonyx jubatus}) is an excellent example of an animal that exhibits complex locomotion. Yet, collecting such complex motion data in laboratory settings, or with traditional motion-capture marker-based systems, is difficult. For instance, GPS-IMU collars treat the animal as a simple point, and harness-based systems are too invasive~\cite{wilson2013locomotion, patel2017tracking}. 

\begin{figure}[t]
	\centering
	\includegraphics[width=\linewidth]{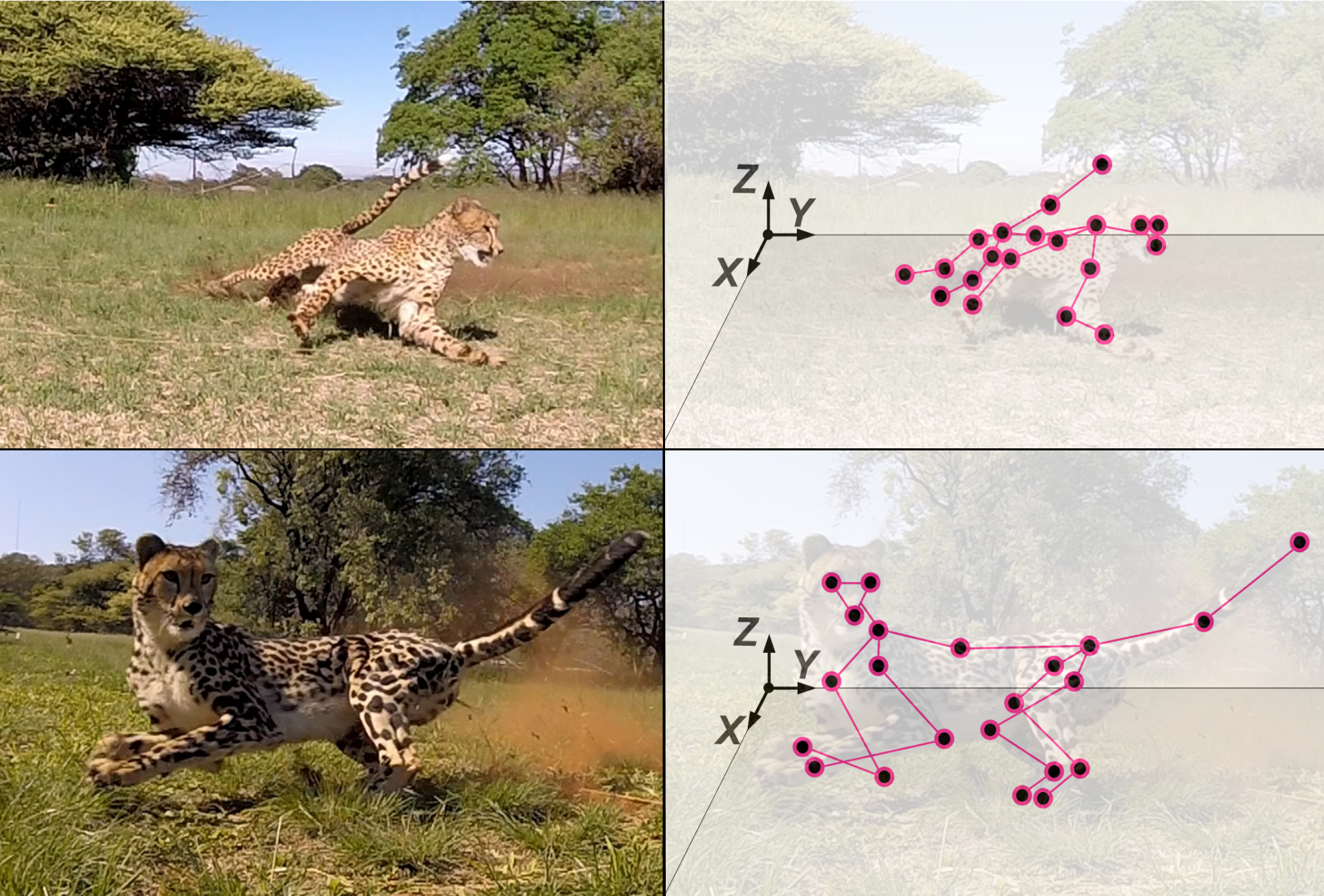}
	\caption{Example frames from the \textit{AcinoSet} dataset alongside 3D reconstruction. The dataset includes frames from various postures, angles, times of the day (and seasons) from 10 different cheetahs. 
}
    \vspace{-5pt}
	\label{fig:acinonetImages} 
\end{figure}

Here, we provide a new dataset of cheetahs running ``in the wild'' named \textit{AcinoSet}, which stems from the scientific name of the cheetah. AcinoSet consists high-speed videos taken from six cameras, camera calibration data and over $7,500$ hand-labeled 2D key points.
Using this dataset we compare three methods of markerless 3D reconstruction: (1) multi-view triangulation~\cite{hartley2003multiple, nath2019using}; (2) an Extended Kalman Filter (EKF)~\cite{patel2017tracking};  and (3) a Full Trajectory Estimation (FTE) method, which is inspired by Moving Horizon Estimation typically used in feedback control~\cite{rao2003constrained} that outperforms the other methods. We believe this 3D dataset, and our strong baseline FTE method, will serve both the robotic and computer vision community as a new 3D benchmark on this highly articulated animal.

\section{Related Work}

In recent years several animal datasets have become available~\cite{Mathis2020APO}. Cao et al. provided around 4k images of several domestic quadrupeds~\cite{Cao_2019_ICCV}. The MacaquePose dataset provides over 13k annotated images of macaques moving from multiple zoos and other sources~\cite{labuguen2020macaquepose}. Biggs et al. provide 20K 2D dog postures based on the  Stanford Dog Dataset~\cite{biggs2020left}. AnimalWeb provides 21K face images~\cite{khan2020animalweb}. The Horse10 dataset comprises 8K horses~\cite{Mathis2019PretrainingBO}. All these datasets are 2D datasets. OpenMonkeyStudio provided an excellent dataset of 195,228 frames with multiple cameras, but in a ``green-studio'' laboratory condition~\cite{bala2020automated}. We previously released a set of about $900$ Cheetah images in-the-wild~\cite{nath2019using} but none aimed towards a 3D benchmark animal pose in-the-wild benchmark, as we do in this paper.

Methods for 3D markerless motion capture of humans are benchmarked on large-scale labelled datasets, such as Human 3.6M~\cite{ionescu2013human3}, HumanEva~\cite{sigal2010humaneva},  NBA2K~\cite{zhu2020reconstructing} or AMASS~\cite{mahmood2019amass}. Methods (surveyed in~\cite{zheng2020deep}) can be divided into model-based~\cite{loper2015smpl,huang2017towards} and model-free, which in turn can be divided into lifting~\cite{mehta2020xnect} and multi-view~\cite{burenius20133d,belagiannis20143d,rhodin2018unsupervised,arnab2019exploiting,qiu2019cross,yang20183d,chen2019unsupervised,yao2019monet}. Thereby, the majority of state-of-the-art methods rely on deep learning (see Zheng et al. for an excellent survey~\cite{zheng2020deep}). Markerless 3D animal pose estimation has previously been done using photogrammetry \cite{sellers2014markerless} but this method requires strong lighting and background texture. An exciting approach is the SMAL method~\cite{loper2015smpl,huang2017towards}, which leverages the 3D scans of toys of animals to obtain 3D shape and texture~\cite{zuffi2018lions, zuffi2019three,biggs2020left}. Additionally, 2D/3D pose estimation of non-human animals has been used for biomechanical, neuroscience, and other applications~\cite{mathis2018deeplabcut, biggs2018creatures, graving2019deepposekit, yao2019monet}. To advance 3D methods for pose estimation on animals, we provide a benchmark and present three baseline 3D methods.

\section{Methods and Data}

\subsection{The AcinoSet Dataset}
\subsubsection{\textbf{Video Collection}}
\begin{figure}[ht]
	\centering
    \input{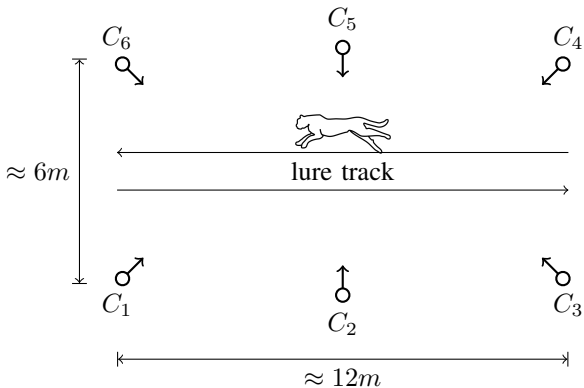}
    \vspace{-3pt}
	\caption{Camera layout for observing cheetah enrichment exercise runs.}

\label{fig:cameraLAyout} 
\end{figure}

Footage of 10 cheetahs was captured during enrichment exercises at the Ann van Dyk Centre (Hartbeespoort, South Africa) and Cheetah Outreach (Somerset West, South Africa) in 2017 and 2019. The footage was captured using six GoPro cameras: either a set of Hero 7 Black cameras at a resolution of $2704\times1520$ at $120$ frames per second (fps) or a set of Hero 5 Session cameras at a resolution of $1920\times1080$ at $90$ frames per second (fps), in the configuration depicted in Fig.~\ref{fig:cameraLAyout}. In total $93$ video sequences were collected, consisting of $42$ runs (straight galloping) and $51$ maneuvers (turning and acceleration). Time synchronization information was included in the fields of view of the cameras by flashing a custom LED rig three times at the beginning and end of each experiment. Calibration (intrinsic and extrinsic) was done using a combination of MATLAB camera calibration toolbox and OpenCV~\cite{opencv_library} via Sparse Bundle Adjustment~\cite{hartley2003multiple}. Video collection was approved by the University of Cape Town Science Faculty Animal Ethics Committee.
\vspace{-10pt}

\subsubsection{\textbf{Data Curation}}

\begin{figure}[ht]
    \centering
    \includegraphics[scale=0.205]{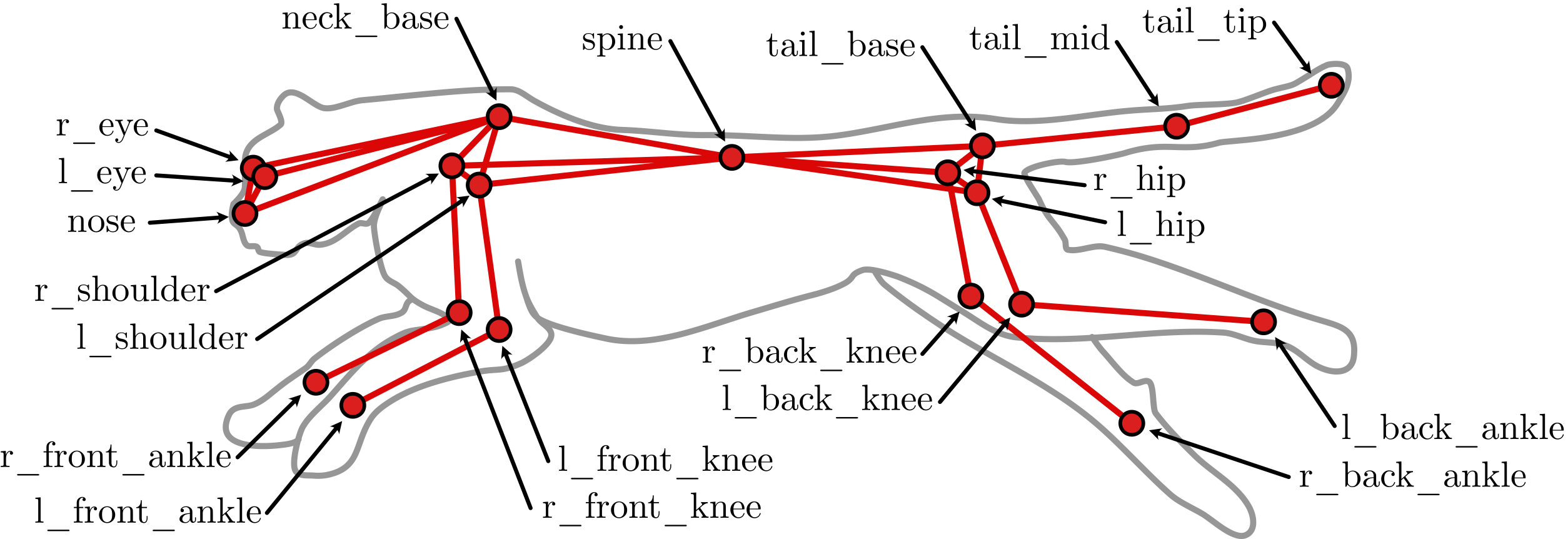}
    \caption{Cheetah rigid body model with markers locations depicted. The lengths of the links were obtained by a sample of measurements taken from several cheetahs at Ann van Dyk Cheetah Centre (South Africa).}
    \label{fig:cheetah_model_plot}
\end{figure}

\begin{figure}[b]
    \centering
    \includegraphics[width=\linewidth]{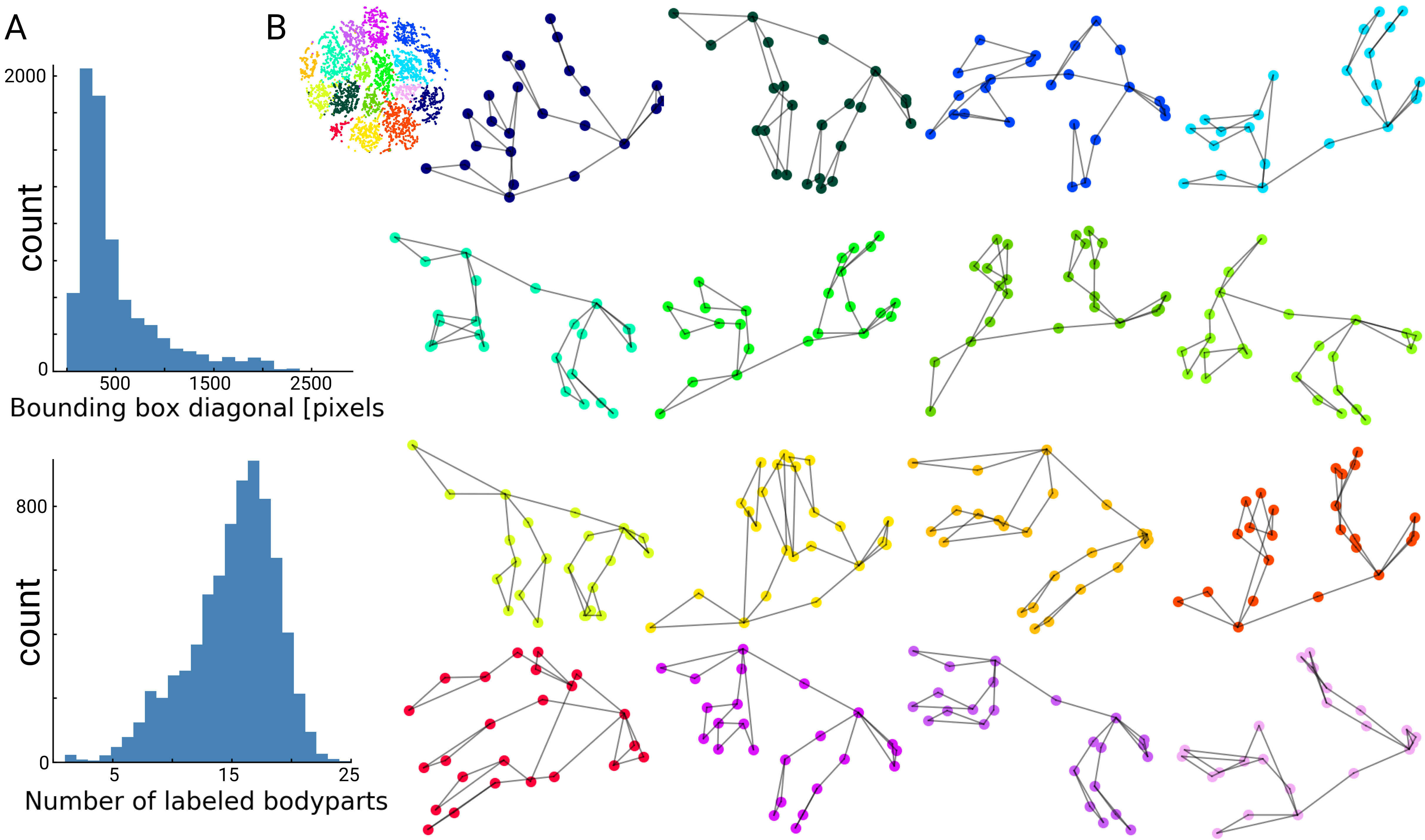}
    \caption{Dataset statistics A: Histogram of number of key points per annotated image, and range of cheetah sizes. B: Diversity of cheetah poses in annotated data seen by t-SNE clustering of the ground truth data, with example centroid posture from each kmeans cluster (cheetahs normalized for size and clustered by kmeans).}
    \label{fig:clusters}
\end{figure}

From these 10 Cheetahs, $7588$ images ($\sim$7\% of the total video frames), were selected (from a uniform random distribution) and 20 key points were expertly annotated (Fig.~\ref{fig:cheetah_model_plot}). Frames from multiple cheetahs (Table~\ref{tab:2d}) consisted of a selection of poses within the cheetahs galloping gait as well transient maneuvers (turning, acceleration and braking, see Fig.~\ref{fig:clusters}). Note that the videos have diverse backgrounds, weather, and lighting (Fig.~\ref{fig:acinonetImages}).

\begin{table}[h]
\centering
\caption{2D Dataset}
\begin{tabular}{lllll}
Cheetah ID & Videos  & Days & Total Frames & Human Labeled \\
\hline
Menya & 18 & 2 & 5,811 & 853\\
Jules & 117 & 8 & 24,853 & 1450\\
Phantom & 141 & 8 &  38,640 & 1372\\
Lily & 57 & 4 & 12,525 & 1535\\
Cetane & 27 & 2 & 8,547 & 404\\
Kiara & 12 & 1 & 2,910 & 94\\
Romeo & 12 & 1 & 2,946 & 37\\
Zorro & 90 & 5 & 21,259 & 364\\
BigGirl & 8 & 1 & 1,396 & 266\\
Ebony & 3 & 1 & 603 & 155\\
\end{tabular}
\label{tab:2d}
\vspace{-10pt}
\end{table}

\subsection{Training 2D Feature Detectors}

We utilized an ImageNet pre-trained version of ResNet152, modified within the DeepLabCut 2.0 framework~\cite{insafutdinov2016deepercut, mathis2018deeplabcut, nath2019using}. We also configured the network to use pairwise correspondence~\cite{insafutdinov2016deepercut}, which improved the accuracy of the predictions. Data was split into 95\% training images, and 5\% for testing and trained for $1E6$ iterations with the stochastic gradient descent (SGD) optimizer. The RMSE for test images was $9.09 \pm 33.08$ pixels.
The RMSE for test images for a network trained without pairwise correspondence was $15.5 \pm 68.82$ pixels. 
There were still large deviations in RMSE (some with an error of over $2000$ pixels) which caused the large standard deviation values. To remove outliers, we fit a Gaussian distribution model. Outliers could roughly be predicted as points deviating by more than $15$ pixels ($3\sigma$).  Then, we obtained a mean of $1.12$ pixels and standard deviation of $5.02$ pixels.
For computing a normalized RMSE (NRMSE) we calculated the bounding box of the cheetah and divided the RMSE by the square root of (height x width).

\subsection{Cheetah Skeletal Model}
\label{sec:cheetahskelmod}
We present several baseline methods for generating the 3D dataset. For the EKF and the FTE methods, we used a rigid body model of the cheetah (Fig.~\ref{fig:cheetah_model_plot}). A set of parameters, shown in Table~\ref{tab:cheetah_model_parameters}, was used to determine the pose of the cheetah model. The model was defined in terms of a global position and orientation of the cheetah and a set of angles describing the relative rotations between the rigid bodies in the model, totalling 24 generalized coordinates.

\begin{table}[ht]
\caption{Pose parameters for the cheetah model}
\begin{tabular}{ll}
Pose parameters & Description  \\ \hline
$x,~y,~z$ &  Head position in inertial frame\\
$\phi_1,~\theta_1,~\psi_1$ &  Head roll, pitch and yaw in the inertial frame\\
$\phi_2,~\theta_2,~\psi_2$ &  Neck\_base roll, pitch and yaw relative to the head\\
$\theta_3$ &  Front torso pitch relative to the neck\\
$\theta_4,~\psi_4$ &  Back torso pitch and yaw relative to the front torso\\
$\theta_5,~\psi_5$ &  Tail\_base pitch and yaw relative to the back torso\\
$\theta_6,~\psi_6$ &  Tail\_mid pitch and yaw relative to the tail base\\
$\theta_7$ &  L\_shoulder pitch relative to the front torso\\
$\theta_8$ &  L\_front\_knee pitch relative to the l\_shoulder\\
$\theta_9$ &  R\_shoulder pitch relative to the front torso\\
$\theta_{10}$ &  R\_front\_knee pitch relative to the r\_shoulder\\
$\theta_{11}$ &  L\_hip pitch relative to the back torso\\
$\theta_{12}$ &  L\_back\_knee pitch relative to the l\_hip\\
$\theta_{13}$ &  R\_hip pitch relative to the back torso\\
$\theta_{14}$ &  R\_back\_knee pitch relative to the r\_hip\\
\end{tabular}
\vspace{-5pt}
\label{tab:cheetah_model_parameters}
\end{table}

Rotation matrices were defined to relate the orientations of the rigid links in the model; these rotation matrices were defined as coordinate rotations according to~\cite{diebel2006representing}. In order to avoid singularities, the order of rotations was chosen such that the second rotation would be less than $90^\circ$~\cite{diebel2006representing}. Using rigid body kinematics, the 3D positions of the markers could be obtained using the lengths and pose parameters~\cite{patel2017tracking}.

The positions in the model were mostly based on the predictions from the DeepLabCut model; however, the paws were excluded as the grass occluded them in most of the videos (the ankles are still captured, which allows for a good 3D representation of locomotion). A ``head'' position, located directly between the eyes of the cheetah, was added to define the cheetah's position in the inertial frame. The front torso was defined as the rigid body containing the markers neck base, spine and shoulder. Similarly the back torso was comprised of the spine, tail base and hip markers. Measurements of the limbs of a subset of cheetahs were obtained from the Ann van Dyk Cheetah Centre and used to derive the positions as described in Table \ref{tab:cheetah_model_positions}. These vary between cheetahs but we found the measurements to be reasonable approximations for the cheetahs.

\begin{table}[h]
\centering
\caption{Positions for the kinematic model of the cheetah}
\begin{tabular}{l}

$\mathbf{p}_{head}=\begin{bmatrix}x& y& z\end{bmatrix}^T$\\
$\mathbf{p}_{l\_eye}=\mathbf{p}_{head} + \mathbf{R}_i^{1} \begin{bmatrix}0& 0.03& 0\end{bmatrix}^T$\\
$\mathbf{p}_{r\_eye}=\mathbf{p}_{head} + \mathbf{R}_i^{1} \begin{bmatrix}0& -0.03& 0\end{bmatrix}^T$\\
$\mathbf{p}_{nose}=\mathbf{p}_{head} + \mathbf{R}_i^{1} \begin{bmatrix}0.055& 0& -0.055\end{bmatrix}^T$\\

$\mathbf{p}_{neck\_base}=\mathbf{p}_{head} + \mathbf{R}_i^{2} \begin{bmatrix}-0.28& 0& 0\end{bmatrix}^T$\\
$\mathbf{p}_{spine}=\mathbf{p}_{neck\_base} + \mathbf{R}_i^{3} \begin{bmatrix}-0.37& 0& 0\end{bmatrix}^T$\\
$\mathbf{p}_{tail\_base}=\mathbf{p}_{spine} + \mathbf{R}_i^{4} \begin{bmatrix}-0.37& 0& 0\end{bmatrix}^T$\\
$\mathbf{p}_{tail\_mid}=\mathbf{p}_{tail\_base} + \mathbf{R}_i^{5} \begin{bmatrix}-0.28& 0& 0\end{bmatrix}^T$\\
$\mathbf{p}_{tail\_tip}=\mathbf{p}_{tail\_mid} + \mathbf{R}_i^{6} \begin{bmatrix}-0.36& 0& 0\end{bmatrix}^T$\\

$\mathbf{p}_{l\_shoulder}=\mathbf{p}_{neck\_base} + \mathbf{R}_i^{3} \begin{bmatrix}-0.04& 0.08& -0.10\end{bmatrix}^T$\\
$\mathbf{p}_{l\_front\_knee}=\mathbf{p}_{l\_shoulder} + \mathbf{R}_i^{7} \begin{bmatrix}0& 0& -0.24\end{bmatrix}^T$\\
$\mathbf{p}_{l\_front\_ankle}=\mathbf{p}_{l\_front\_knee} + \mathbf{R}_i^{8} \begin{bmatrix}0& 0& -0.28\end{bmatrix}^T$\\

$\mathbf{p}_{r\_shoulder}=\mathbf{p}_{neck\_base} + \mathbf{R}_i^{3} \begin{bmatrix}-0.04& -0.08& -0.10\end{bmatrix}^T$\\
$\mathbf{p}_{r\_front\_knee}=\mathbf{p}_{r\_shoulder} + \mathbf{R}_i^{9} \begin{bmatrix}0& 0& -0.24\end{bmatrix}^T$\\
$\mathbf{p}_{r\_front\_ankle}=\mathbf{p}_{r\_front\_knee} + \mathbf{R}_i^{10} \begin{bmatrix}0& 0& -0.28\end{bmatrix}^T$\\

$\mathbf{p}_{l\_hip}=\mathbf{p}_{tail\_base} + \mathbf{R}_i^{4} \begin{bmatrix}0.12& 0.08& -0.06\end{bmatrix}^T$\\
$\mathbf{p}_{l\_back\_knee}=\mathbf{p}_{l\_hip} + \mathbf{R}_i^{11} \begin{bmatrix}0& 0& -0.32\end{bmatrix}^T$\\
$\mathbf{p}_{l\_back\_ankle}=\mathbf{p}_{l\_back\_knee} + \mathbf{R}_i^{12} \begin{bmatrix}0& 0& -0.25\end{bmatrix}^T$\\

$\mathbf{p}_{r\_hip}=\mathbf{p}_{tail\_base} + \mathbf{R}_i^{4} \begin{bmatrix}0.12& 0.08& -0.06\end{bmatrix}^T$\\
$\mathbf{p}_{r\_back\_knee}=\mathbf{p}_{r\_hip} + \mathbf{R}_i^{13} \begin{bmatrix}0& 0& -0.32\end{bmatrix}^T$\\
$\mathbf{p}_{r\_back\_ankle}=\mathbf{p}_{r\_back\_knee} + \mathbf{R}_i^{14} \begin{bmatrix}0& 0& -0.25\end{bmatrix}^T$\\

\end{tabular}

\label{tab:cheetah_model_positions}
\end{table}

\section{Results}

The main contributions of this work are three-fold: 
\begin{enumerate}
    \item  The large 2D dataset ($>$100K frames, with 7588 human annotated frames, as described in Methods \& Data).
    \item Three baseline methods for generated the 3D data, including FTE, which provides excellent results.
    \item Human-validated 3D ground-truth frames and graphical user interface tools to validate more. 
\end{enumerate}

\begin{figure*}
    \centering
    \includegraphics[scale=0.66]{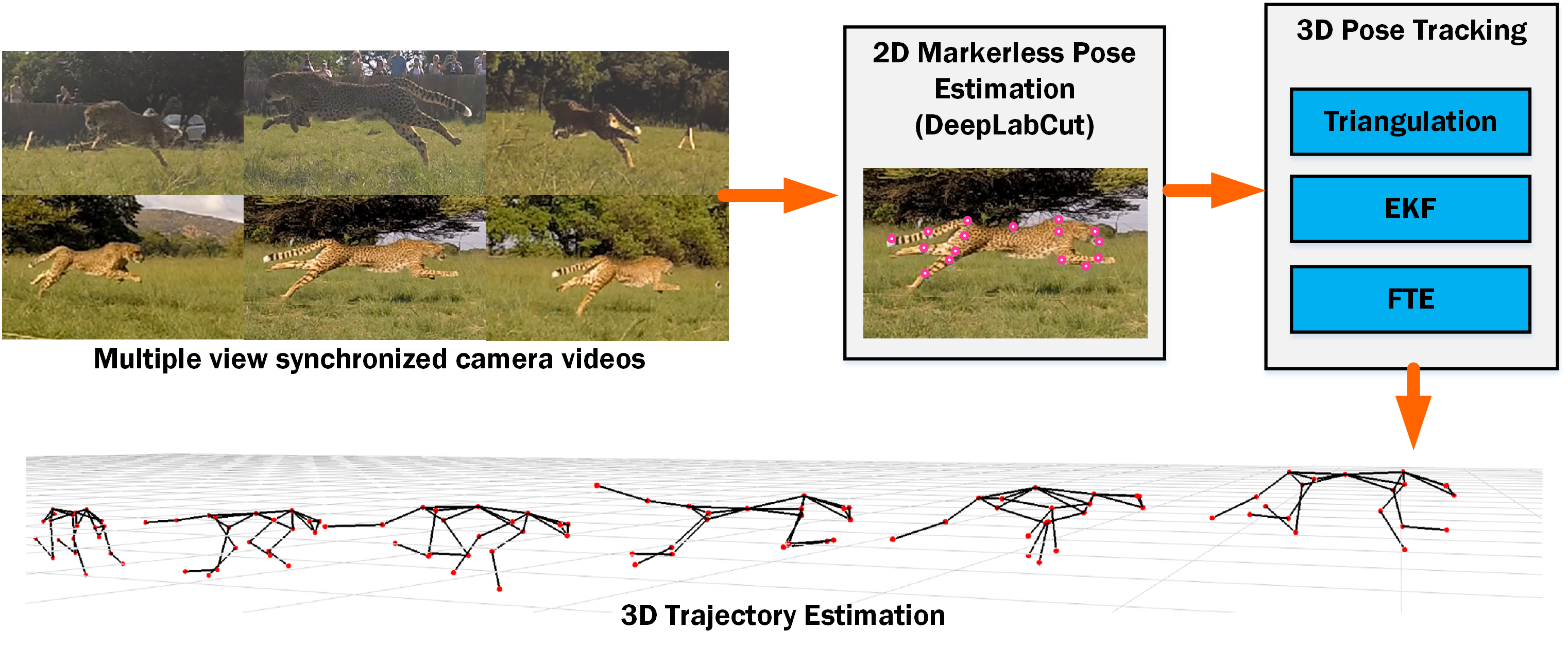}
    \caption{\textbf{An overview of the 3D pose tracking}. Multi-view videos are fed into a markerless pose estimation algorithm (DeepLabCut) which outputs 2D pose estimates in each view. These views are then combined using one of the three 3D pose estimation methods (Triangulation, EKF or FTE).}
    \label{fig:overallAlgorithm}
    \vspace{-10pt}
\end{figure*}

To create a strong baseline 3D dataset, we used three methods (Fig.~\ref{fig:overallAlgorithm}) to compute 3D skeletons, which utilized the 2D key point estimates from the $119,490$ video frames.
\vspace{-5pt}

\subsection{Triangulation} 

Our first baseline is sparse bundle adjustment~\cite{hartley2003multiple} using SciPy~\cite{virtanen2020scipy}. Unlike in our prior work~\cite{nath2019using}, we used a Cauchy robust cost function to prevent outliers from skewing the 3D point estimates to the naive squared error approach~\cite{hartley2003multiple}. The initial 3D point estimates were obtained for 2D correspondences with a likelihood above $0.5$. 

\vspace{-5pt}
\subsection{Extended Kalman Filter}
Our second baseline is another popular method for multi-camera 3D reconstruction, namely the Extended Kalman Filter (EKF)~\cite{forsyth2002computer}. In our configuration the EKF utilizes the rigid body model described in Section~\ref{sec:cheetahskelmod} and factors in the covariances (measurement and model) to increase the accuracy of the state estimates.

\subsubsection{Process Model}
The skeletal motion was assumed to behave according to a constant acceleration model (with sample time $\Delta t$). The kinematic state vector $\hat{\mathbf{x}}$ therefore consisted of the pose parameters $\mathbf{q}$ (Table \ref{tab:cheetah_model_parameters}), their velocities $\mathbf{\dot{q}}$ and their accelerations $\mathbf{\ddot{q}}$. Jerk in the pose parameters $i$ was accounted for as a tunable covariance ($\sigma_i$) for smoothness\cite{labbe2014}.

\subsubsection{Measurement Model}
A measurement function for the EKF was required to produce a set of measurement estimates given a set of state estimates $\hat{\mathbf{x}}$. The 3D positions of each marker for the cheetah could be determined using the 3D kinematic equations. These 3D positions were then projected into the view of each camera taking into account the fisheye distortion effects~\cite{patel2017tracking}. Denoting the estimated measurement for coordinate $c$ (either $u$ or $v$) for a marker $m$ seen by camera $i$ as $\hat{\mathbf{z}}_{i, m, c}$, the measurement equation provides a measurement estimate $\hat{\mathbf{z}}$ comprising 240 elements (6 cameras $\times$ 20 markers $\times$ 2 pixel coordinates):
\vspace{-1pt}
\begin{equation}
    \hat{\mathbf{z}} = h(\hat{\mathbf{x}}) =
    \begin{bmatrix}
        \hat{\mathbf{z}}_{1, l\_eye, u}\\
        \hat{\mathbf{z}}_{1, l\_eye, v}\\
        \vdots\\
        \hat{\mathbf{z}}_{1, r\_back\_ankle, v}\\
        \hat{\mathbf{z}}_{2, l\_eye, u}\\
        \vdots\\
        \hat{\mathbf{z}}_{6, r\_back\_ankle, v}\\
    \end{bmatrix}.
    \label{eqn:ekf_measurement_estimate}
\end{equation}

The measurement covariance matrix $\mathbf{R}$ was constructed as a diagonal matrix where each diagonal element was equal to the measurement variance from DeepLabCut ($\sigma^2 = (5\text{ pixels})^2$) for high likelihoods and 
$2704^2$ for low likelihoods, which is the square of the maximum possible pixel error when a cheetah was in the frame: $\mathbf{R} = I \sigma^2.$ 

\subsubsection{Outlier Rejection}
In order to prevent outliers from causing the state to diverge, an extra step was added to the update stage of the EKF: if an outlier was detected, the innovation (residual) for that measurement was set to zero. An outlier could be determined using the innovation, $\Tilde{\mathbf{y}}$, and innovation covariance, $\mathbf{S}$. Outliers were assumed to be measurements whose innovation was three times greater than the square root of the innovation covariance~\cite{fang2018robustifying}. 

\subsection{Full Trajectory Estimation}

One downside of the EKF is that estimates only depend on the previous sample and can diverge if multiple outliers are present in time. Thus, we consider the entire trajectory of states and measurements simultaneously. This also allows us to impose state constraints which is not possible in the EKF, akin to Moving Horizon Estimation (MHE)~\cite{rao2003constrained}, yet here we optimize over the full trajectory.

\subsubsection{Parameters and variables}
A trajectory, $\mathbf{x}$, consists of a set of pose parameters at each time step. These pose parameters are the positions and angles, $[x,y,z,\phi_1 \dots \theta_{14}]$, which allow us to define the 3D marker positions of the cheetah.
The first and second order time derivatives of $\mathbf{x}$, namely $\dot{\mathbf{x}}$ and $\ddot{\mathbf{x}}$, describe the ``velocities'' and ``accelerations'' of the pose parameters respectively. An array, $\mathbf{s}$, contained the 3D marker positions for each time step. The complete list of variables are listed in Table~\ref{tab:traj_vars}. An array, $\mathbf{y}$, contained 2D keypoints with shape $N\times c \times m \times 2$, with $N$ the number of time steps in the trajectory, $p$ the number of pose parameters ($24$), $c$ the  number of cameras ($6$) and $m$ the number of markers on the cheetah ($20$).  

\begin{table}[ht]
\centering
\caption{Trajectory optimization variables.}
\begin{tabular}{lll}
Variables & Length & Description \\
\hline
$\mathbf{x}$ & $N\times p$ & Pose parameters across trajectory\\
$\dot{\mathbf{x}}$ & $N\times p$ & Time derivative of $\mathbf{x}$\\
$\ddot{\mathbf{x}}$ & $N\times p$ & Time derivative of $\dot{\mathbf{x}}$\\
$\mathbf{s}$ & $N \times m \times 3$ & 3D marker positions\\
$\mathbf{w}$ & $N\times p-1$ & Model noise\\
$\mathbf{v}$ & $N\times c \times m \times 2$ & Measurement noise\\
\end{tabular}

\label{tab:traj_vars}
\end{table}

\subsubsection{Modeling Constraints}
Similar to the EKF, the model constraints between time steps used a constant acceleration model. In reality, the pose parameters did not obey this model and so $\mathbf{w}$ was added to account for the acceleration errors (or \textit{disturbances}) for each pose parameter between time steps. Implicit Euler integration was used to formulate the equations below:
\vspace{-5pt}
\begin{align}
        \mathbf{x}_k &= \mathbf{x}_{k-1} + \Delta t\dot{\mathbf{x}}_{k}
        \quad\text{for}\quad k=\{2,\dots,N\}
    \label{eqn:traj_constraint_pos} \\
    \dot{\mathbf{x}}_k &= \dot{\mathbf{x}}_{k-1} + \Delta t\ddot{\mathbf{x}}_{k}
    \quad\text{for}\quad k=\{2,\dots,N\}
    \label{eqn:traj_constraint_vel} \\
    \ddot{\mathbf{x}}_k &= \ddot{\mathbf{x}}_{k-1} + \mathbf{w}_k
    \quad\text{for}\quad k=\{2,\dots,N\}.
    \label{eqn:traj_constraint_acc}
\end{align}
In addition to the constraints for the motion model, bound constraints on the pose parameters were added to ensure physically plausible motion.

\subsubsection{Measurement constraints}
The estimated pose parameters need to be related to the measurements in order to determine how well the estimated pose relates to what is seen by the cameras. In order to simplify the measurement estimate constraints, the problem was split into two parts. First, a constraint was set up that relates the pose parameters at each time step to a set of 3D marker positions through a function, $f$:
\begin{equation}
    \mathbf{s}_k = f(\mathbf{x}_k)
    \quad\text{for}\quad k=\{2,\dots,N\}.
    \label{eqn:traj_constraint_pose}
\end{equation}

A second constraint was set up which projected these 3D marker positions to 2D measurement estimates, $\hat{\mathbf{y}}$, for each camera and equated the measurement estimates to the measurements, $\mathbf{y}$, from DeepLabCut. A variable $\mathbf{v}$ was added to account for measurement noise or \textit{measurement disturbances}. A projection function, $h$, for each camera was used to project the 3D points in order to obtain the 2D measurement estimates:
\begin{multline}
    \mathbf{y}_{i, j} = h_j(\mathbf{s}_{i, j}) + \mathbf{v}_{i, j}
    \quad\text{for}\quad \\
    i=\{1,\dots,n\}, \quad j=\{1,\dots,c\}.
    \label{eqn:traj_constraint_meas}
\end{multline}

\subsubsection{Cost function}
Given the constraints above, we have a set of measurement errors, $\mathbf{v}$, and acceleration errors, $\mathbf{w}$, which we would like to minimize. The measurement error, $e_{\text{\it meas}}$, can be formulated in a way that is very similar to that of the triangulation. In order to reduce the effect of outliers, the re-descending robust cost function $C(\centerdot)$~\cite{nicholson2014line} with values $a=3, b=10, c=20$ was used.

The measurement costs were normalised by dividing by the standard deviation of the measurements, $\sigma_{\text{\it meas}}=5$ pixels:
\begin{equation}
    e_{\text{\it meas}} = \sum_{i=1}^n \sum_{j=1}^c \sum_{k=1}^m \sum_{l=1}^2 C\bigg(\frac{\mathbf{v}_{i,j,k,l}}{\sigma_{\text{\it meas}}}\bigg).
    \label{eqn:traj_cost_meas}
\end{equation}

The model error, $e_{\text{\it model}}$, was the sum of squares of acceleration noise. In order to normalise this noise, the squared acceleration error values were divided by the acceleration variance of the pose parameter they correspond to. Alternatively, the normalisation can be thought of as dividing by the standard deviation before squaring the value:
\begin{equation}
    e_{\text{\it model}} = \sum_{i=1}^n \sum_{j=1}^p \frac{\mathbf{w}_{i,j}^2}{{\sigma_{\text{\it model}}^2}_j} = \sum_{i=1}^n \sum_{j=1}^p \bigg(\frac{\mathbf{w}_{i,j}}{{\sigma_{\text{\it model}}}_j}\bigg)^2.
    \label{eqn:traj_cost_model}
\end{equation}
The minimization problem could then simply be formulated as per \eqref{eqn:traj_est_min} subject to all of the model and measurement constraints:
\vspace{-2pt}
\begin{equation}
    \underset{\text{$\mathbf{x}$, $\dot{\mathbf{x}}$, $\ddot{\mathbf{x}}$, $\mathbf{s}$, $\mathbf{w}$, $\mathbf{v}$}}{\text{min}} \quad e_{\text{\it meas}} + e_{\text{\it model}}.
    \label{eqn:traj_est_min}
\end{equation}

\subsubsection{Implementation}
The optimization was implemented using Pyomo~\cite{hart2017pyomo} and IPOPT~\cite{wachter2006implementation} with the MA86 linear solver~\cite{hsl2007collection}. Similarly to the EKF, for each run the position and orientation of the cheetah was estimated using the triangulated points. All of the other states were initialised to zero (so that the cheetah was in a valid, upright pose).

\section{3D Baseline Model Testing}


First, to evaluate each of the 3D reconstruction techniques, we constructed synthetic ground truth data (to mitigate any errors in feature detection, camera calibration, or human-annotation). Thus, a cheetah run was simulated by defining a 3D pose (using the intrinsic and extrinsic parameters we measured), corresponding to marker positions, for a cheetah as a function of pose parameters. A 3D trajectory was then created by adjusting the pose parameters at each frame to create a realistic-looking cheetah run.

Several simulated datasets were used for testing and evaluating the different trajectory estimation methods; specifically, datasets which allowed for the evaluation of these methods in the presence of noise and outliers in the measurement data (the 2D points seen by the cameras). Varying degrees of Gaussian noise were added to the ground truth 2D points. The noise $n$ for each 2D coordinate was randomly drawn from a Gaussian distribution with $0$ mean and standard deviation $\sigma_{n}$, so $n \sim \mathcal{N}(0, \sigma_{n}^2)$.

Outliers were created by adding an ``outlier value'' $o$ to a randomly selected subset of the ground truth 2D points. The probability of a 2D point being turned into an outlier was determined by $p_o$. The outlier value $o$ for each 2D coordinate was randomly drawn from a Gaussian distribution with 0 mean and standard deviation $\sigma_{o}$, so $o \sim \mathcal{N}(0, \sigma_{o}^2)$.

Each modified simulated 2D point coordinate $c_\text{\it{noisy}}$ was obtained using
\begin{equation}
  c_\text{\it{noisy}} = \begin{cases}
  c + n \quad & \text{for noise only} \\
  c + n + o \quad & \text{for outliers (including noise)}
  \end{cases}
\end{equation}
where $c$ is one of the $(u, v)$ pixel coordinates.

\begin{figure}[ht]
    \centering
    \includegraphics[width=.78\linewidth]{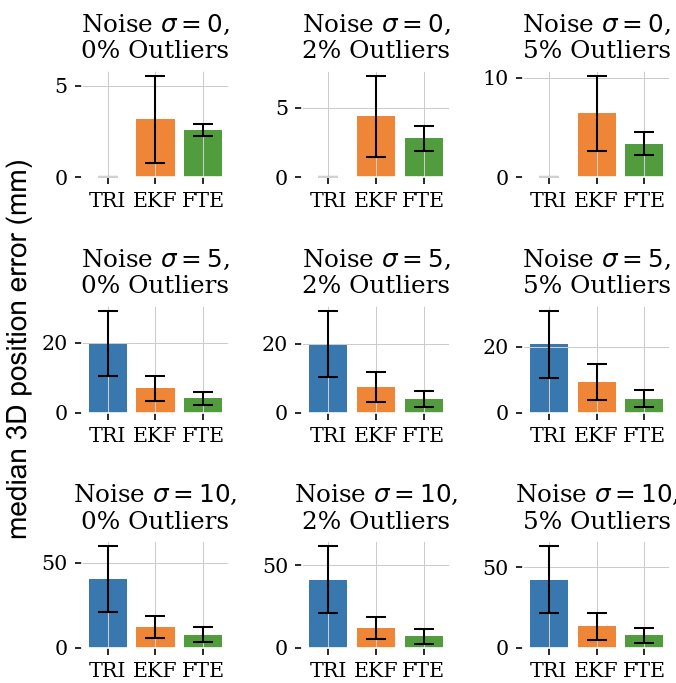}
    \caption{Summary of simulation results of the three methods.}
    \vspace{-10pt}
    \label{fig:simulation_results_plot}
\end{figure}

The results for all three methods are summarised in Fig.~\ref{fig:simulation_results_plot}, which compares the median absolute difference and median. 

\vspace{-5pt}
\subsection{Baseline 3D on AcinoSet}
All three methods were next validated on the \textit{AcinoSet} (Fig.~\ref{fig:dive_quality_comarison}). A visual comparison of the three methods can also be found in the supplementary video.

To quantitatively evaluate each of the three methods, we reprojected 3D point predictions for the different methods to the camera planes of each of the $6$ cameras. Each of the methods was given filtered multi-view 2D data (points with a confidence value of less than $0.5$ were discarded, as they are often occluded). We then compared these reprojected 2D points with 2D ground truth data for the corresponding frames using the RMSE. Two poses were considered: one simple ``run'' pose with high visibility for all body parts and minimal occlusion, and one more complicated ``dive'' pose with low visibility and high levels of occlusion (see Fig. \ref{fig:dive_quality_comarison}). For each pose, $1000$ hand-checked 3D ground truth points were reprojected to each of the $6$ image planes, for a total of $6000$ 2D ground truth points. The RMSE and standard error of mean (SEM), along with a normalised error (NRMSE) scaling with the size of the cheetah in frame, for each method are shown in Table VII. Consistent with the visual impression from Fig.~\ref{fig:dive_quality_comarison}, full trajectory estimation performed best. 

\begin{table}[ht]
\caption{RMSE, SEM, and NRMSE in pixels for each method}
\centering
\begin{tabular}{ccrrr}
Pose Type & Metric & TRI & EKF & FTE \\
\hline
Run & RMSE & $28.24$ & $3.40$ & \textbf{$2.76$} \\
  & SEM & $0.26$ & $0.03$ & $0.03$ \\
  & NRMSE & $0.17$ & $0.02$ & $0.02$ \\
\hline
Dive & RMSE & $76.35$ & $39.40$ & \textbf{$38.44$} \\
  & SEM & $0.70$ & $0.36$ & $0.35$ \\
  & NRMSE & $0.56$ & $0.29$ & $0.28$ \\
\end{tabular}
\label{tab:rmse_compare}
\end{table}

\begin{figure}
    \centering
    \includegraphics[width=\linewidth]{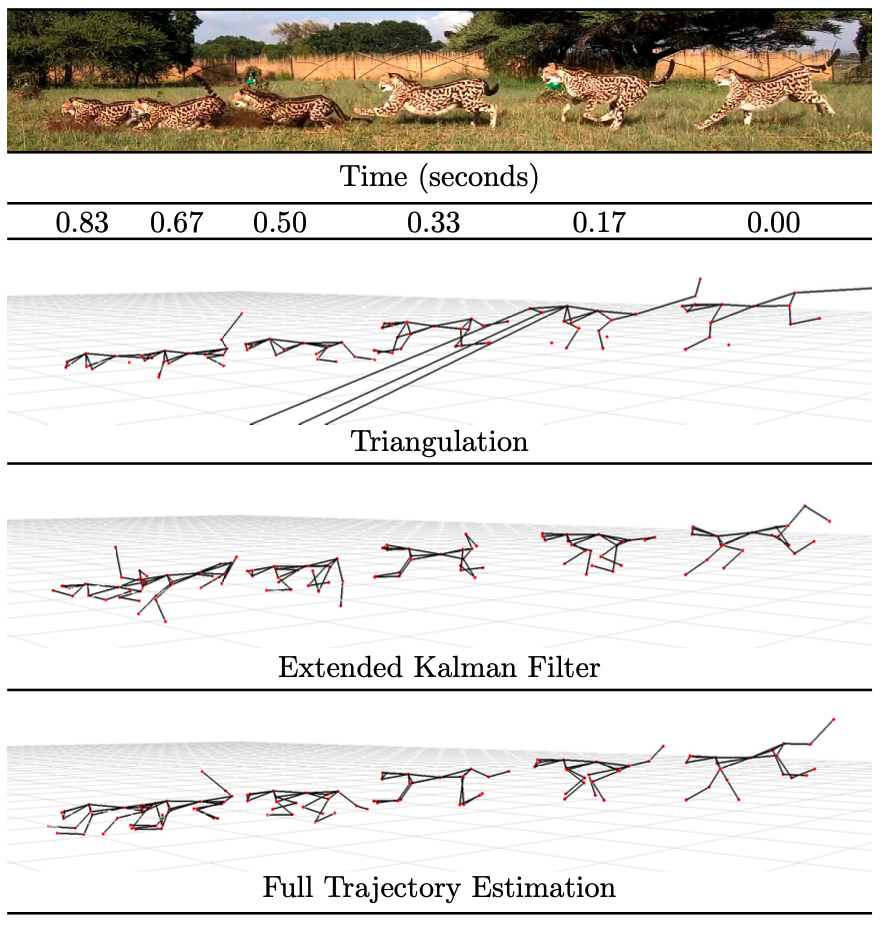}
    \caption{Examples of the 3D baseline models run on real cheetah footage.}
    \label{fig:dive_quality_comarison}
\end{figure}

\vspace{-5pt}
\section{3D Ground Truth Validation}

As the FTE performed the best, we processed all the $>$119K frames to create the \textit{3D AcinoSet}. This results in 19,915 3D frames. 
To check the quality of these 3D postures---as this dataset should serve as a strong benchmark for 3D pose estimation algorithm development---we built a new graphical user interface in order to swiftly check, or correct, 3D ground truth as needed.
We used this tool to estimate the amount of human corrections needed to be ``perfect'' (most needed no corrections, and minor corrections were under 10 pixels, checked on n=600 frames) and the largest adjustments were typically in the extremities, such as the tail and ankles. The proportion of large outliers where the adjustment was over $100$ mm in magnitude was only $0.71\%$.

\section{Discussion and Conclusions}

We present the development of a new animal multi-camera pose estimation dataset, termed \textit{AcinoSet}, for 3D pose estimation. With over $7500$ human-annotated images, plus 6 camera data and calibration information, we believe this dataset will be an attractive platform for benchmarking new 3D pose estimation tools. This dataset may be beneficial for researchers in fields such as biology and robotics: in biology, there is a push to bring biomechanics into the field~\cite{bauer2020mechanical, Hausmann2021MeasuringAM}, and in robotics, for researchers developing deep-learning based controllers~\cite{peng2020learning}.

We also compared three baseline methods for 3D reconstruction. We report that the most robust method, both quantitatively and qualitatively, was the FTE method. Unsurprisingly, this and the EKF outperformed triangulation, likely as they also model the temporal continuity at the cost of  greater computational complexity. 

The two main advantages of FTE are that the pose parameters can be constrained to their natural ranges, and the 3D marker positions constrained to the skeletal model of the cheetah. It does however require a good initial seed, perhaps provided by either the triangulation or the EKF as a warm-start. The EKF method provided a large improvement over the triangulation with the inclusion of temporal information (via states) and a skeletal model. The rigid body model applied is by no means a realistic skeletal model of the cheetah, yet it provides enough structure to generate acceptable 3D trajectories. Taken together, we hope this allows the community to tackle new 3D challenges.

\section*{Acknowledgments}
The authors thank An Chi Chen, Alexandra Barry, Bilal Waleed, Inessa Rajah, James Cushway \& Annet George for assistance with video collection, labelling \& NN training. We also thank Ines Everaert of Ann van Dyk Centre \& Liesl Smith of Cheetah Outreach for allowing access to their cheetahs. MWM is the Bertarelli Foundation Chair of Integrative Neuroscience.

\balance
\bibliographystyle{ieeetr}


\end{document}